\documentclass[sigconf,nonacm,balance=true]{acmart}

\usepackage{algorithm}
\usepackage{algorithmic}
\usepackage{array}

\newcommand{\TraceCount}{210,000}
\newcommand{\RepetitionCount}{7}
\newcommand{\NoCheckInvalid}{60.0\%}
\newcommand{\ATRFalseAllow}{0}
\newcommand{\ATRFalseBlock}{0}

\newcommand{\ATRMedian}{10.8}
\newcommand{\ATRPninetyfive}{13.0}
\newcommand{\ATRPninetynine}{20.5}
\newcommand{\ATRSeedMedianMean}{10.77}
\newcommand{\ATRSeedMedianCI}{0.07}
\newcommand{\FullScanMedian}{39.9}
\newcommand{\NoCheckUnsafeRelease}{100.0}
\newcommand{\VersionAbortSafeReplan}{100.0}
\newcommand{\FullScanPremisesPerChange}{6.0}
\newcommand{\ATRPremisesPerChange}{0.6}
\newcommand{\GraphBytes}{7.4}
\newcommand{\PredicateClosure}{2.6}
\newcommand{\ObjectClosure}{6.4}
\newcommand{\SQLitePairs}{7,000}
\newcommand{\SQLiteDoubleCommits}{0}
\newcommand{\SQLiteZeroCommits}{0}
\newcommand{\SmallestReads}{13}
\newcommand{\SmallestSelective}{8.8}
\newcommand{\SmallestFullScan}{12.1}
\newcommand{\LargestReads}{4,093}
\newcommand{\LargestSelective}{9.3}
\newcommand{\LargestFullScan}{2595.9}

\makeatletter
\renewcommand{\@fnsymbol}[1]{%
  \ifcase#1\or\textdagger\or\textasteriskcentered\or\textdaggerdbl
  \or\textsection\or\textparagraph\or\textbardbl\else\@ctrerr\fi}
\makeatother

\begin{document}

\title{From Version Conflicts to Decision Conflicts: Selective Revalidation for Long-Running AI Agents}

% The first three authors share one native acmart affiliation group.
\author{Yongjian Lyu\texorpdfstring{\textsuperscript{1}}{}}
\authornote{\raggedright Corresponding authors: Yongjian Lyu
(\texttt{lvyongjian@huawei.com}) and Yang Ren
(\texttt{renyang1@huawei.com}).}
\author{Yang Ren\texorpdfstring{\textsuperscript{1}}{}}
\authornotemark[1]
\author{Ruofei Lai\texorpdfstring{\textsuperscript{1}}{}}
\affiliation{%
  \institution{\textsuperscript{1}\,Huawei Technologies Co., Ltd.}
  \city{}
  \country{United Kingdom}}

\author{Wenting Liu\texorpdfstring{\textsuperscript{2}}{}}
\affiliation{%
  \institution{\textsuperscript{2}\,Xi'an Jiaotong University}
  \city{Xi'an 710049}
  \country{China}}

\renewcommand{\shortauthors}{Yongjian Lyu et al.}

\begin{abstract}
Long-running AI agents may read state, reason, wait for tools or human
approval, and perform an external action much later. The state that justified
the action can change in the meantime. For example, after an agent proposes an
80 GBP refund under a limit of 100, a customer-name change affects only
presentation metadata, a new
limit of 90 still permits the refund, a limit of 50 invalidates it, and a
refund issued by another worker must prevent a duplicate. Standard optimistic
concurrency control and version checks can detect that previously read state
has changed, but by themselves do not determine whether that change invalidates
the pending action's justification. We call any detected
version change a \emph{version conflict}; when that change invalidates the
action's justification, it is also a \emph{decision conflict}. ATR records the explicit,
executable conditions that justify a pending action and rechecks only the
conditions affected by a change before releasing the external operation. It
can retain the action, refresh non-decisive metadata, require replanning, or
block execution; a target-side transaction or compare-and-set binds checked
state to commit. Across \TraceCount\ controlled executions over 15 mutation
cases, ATR matched every developer-specified outcome with no false allows or
blocks. In ten durable SQLite checkpoint/resume cells, it evaluated 0.6
conditions per change versus 6.0 for FullScan. At \LargestReads\ recorded
reads, ATR took \LargestSelective\,$\mu$s versus
\LargestFullScan\,$\mu$s for FullScan. These deterministic results establish
controlled feasibility, not production generality or automatic extraction of
the required conditions.
\end{abstract}



\maketitle

\hypersetup{%
  pdftitle={From Version Conflicts to Decision Conflicts: Selective Revalidation for Long-Running AI Agents},
  pdfauthor={Yongjian Lyu, Yang Ren, Ruofei Lai, Wenting Liu},
  pdfcreator={LaTeX with acmart}}

\section{Introduction}

AI agents interleave reasoning, retrieval, and tool actions
~\cite{yao2023react}; durable executions may also cross approval pauses,
retries, and resumptions~\cite{mozafari2026semiso}. This separates the time at
which an agent reads state from the time at which it acts on that state.
Consider a refund agent that observes an order as delivered, defective, not
yet refunded, and worth 80 GBP, together with a delegated refund limit of 100.
It concludes that the customer is eligible, proposes the refund, and waits for
approval. Before execution, the customer display name may change; the limit
may fall to 90 or 50; or another worker may issue the refund. Every update can
change a recorded version. The name change and a new limit of 90 preserve the
justification; a limit of 50 and a duplicate refund invalidate it.

Database transactions and optimistic concurrency control (OCC) remain
responsible for target-local concurrency and atomic updates
~\cite{kung1981occ,berenson1995isolation}. They can report that an agent read
version 11 while the resource is now at version 12. They do not know whether
the changed field matters to the refund. Semantic serializability can use
application meaning to admit more database schedules, but it remains a
correctness criterion for transaction execution
~\cite{brayner1999semantic}. An agent may instead reason across databases,
APIs, documents, and long pauses, so holding one transaction open is neither
practical nor sufficient. The layers answer different questions: OCC asks
whether state changed; ATR asks whether the change invalidates the pending
action's justification; the target transaction or compare-and-set (CAS) asks
whether the action can atomically commit against the checked state.

\begin{figure*}[t]
  \centering
  \includegraphics[width=\textwidth]{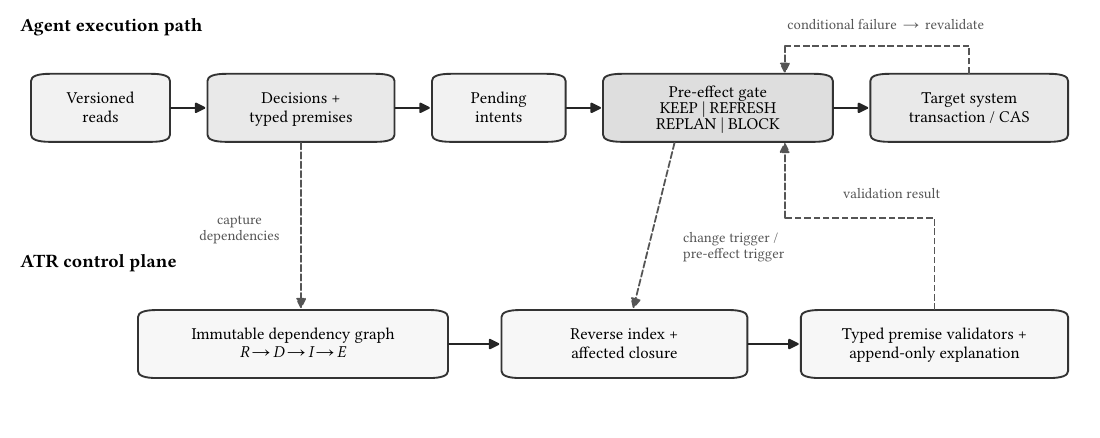}
  \caption{ATR runtime overview. From a changed read, ATR follows recorded
  dependencies to the reachable pending actions (the affected closure) and
  rechecks their required conditions. Permitted operations execute through a
  target-side transaction or CAS; a conditional failure triggers revalidation.
  State held in another system requires a separate binding mechanism.}
  \Description{Versioned observations lead to agent conclusions, proposed
  actions, a pre-action gate, and an authoritative target. An ATR control plane
  contains the dependency graph, reverse index, executable condition checks,
  and explanation log. Solid arrows show
  within-layer progression; dashed arrows show cross-layer triggers and
  validation results; dashed feedback returns conditional failure to the gate.}
  \label{fig:architecture}
\end{figure*}

The key observation is that a version change should trigger inspection, not
automatic invalidation of the agent's decision. ATR records the concrete facts
used by a decision and the executable conditions that must remain true for its
pending action. When related state changes, or immediately before execution,
ATR follows the recorded dependencies and rechecks only the affected
conditions. Figure~\ref{fig:architecture} shows this boundary: ATR decides
whether the proposed action remains justified, while the target database or
API still enforces the atomic operation. This separation avoids claims of
global serializability or exactly-once execution across unrelated systems.

This short paper contributes: (1) a model that distinguishes version conflicts
from decision conflicts; (2) a reverse-index protocol with four operational
outcomes and target-side checked-state binding; and (3) an executable prototype and
controlled evaluation of safety, selectivity, scaling, races, and dependency
omission. Our results support feasibility, not production generality: the
workloads use explicit developer-authored conditions and deterministic
ground truth.

\section{Model and Semantics}

\subsection{Agent Transactions and Premises}

We first ground the terminology in the refund example. \emph{Evidence} is the
concrete state observed by the agent: delivered${}={}$true,
defect${}={}$true, refunded${}={}$false, amount${}={}$80, and
refund-limit${}={}$100. A \emph{decision} is the conclusion supported by those
observations---here, that the order is eligible for an 80 GBP refund. A
\emph{premise} is a condition that must remain true for the decision or action
to stay justified. The refund depends on five premises: the order was
delivered; a defect exists; no refund has been issued; 80 does not exceed the
current limit; and the agent has refund authority. A \emph{typed premise}
represents such a condition with defined fields and semantics so an executable
validator can check it again. Free-form chain-of-thought or rationale may aid
an audit, but is not itself a premise on which ATR relies. The \emph{intent} is
the proposed ``issue an 80 GBP refund'' operation; the \emph{effect} is the
actual external operation that commits or sends the refund.

Formally, an agent transaction is $T=\langle R,D,I,E\rangle$. In the example,
$R$ contains the order and policy observations, $D$ the eligibility decision,
$I$ the proposed refund, and $E$ the attempted or committed refund operation.
Each read $r\in R$ stores a resource key, value digest, connector-defined
version, and read policy. Each decision $d\in D$ stores its outcome and typed
premises. A directed graph $G=(V,\mathcal{A})$ contains these immutable nodes,
with $u\rightarrow v$ meaning that $v$ may depend on $u$. The common path is
$R\rightarrow D\rightarrow I\rightarrow E$, with many-to-many edges. For a
changed read, the \emph{affected closure} is the downstream part of this graph
that can reach a pending action; a reverse index locates its starting nodes.

A premise $p$ records evidence fields $X$, a deterministic validator $f$,
declared failure severity $\mathit{sev}(p)$, and validator/policy version $q$.
Here $\mathit{sev}(p)$ is \textsf{REPLAN} or \textsf{BLOCK}. Provenance
standards provide a vocabulary for entities and derivations
~\cite{moreau2013prov}; ATR adds explicit decision conditions and the gate
before an external operation.

Let $P(e)$ contain all required premises reachable from pending effect $e$; a
premise is required when its failure changes the operational response for the
current intent. A premise is \emph{block-critical} when its failure severity is
\textsf{BLOCK}, or policy marks it as safety-, authority-, compliance-, or
irreversible-effect-critical. Let $B(e)$, the \emph{binding set}, be the
evidence versions or tokens that the target can check atomically when applying
$e$. Given current state
$S_v$ at versions $v$, the effect is \emph{decision-consistent at the gate}
when every required $p\in P(e)$ evaluates true, except that an explicit
refresh-only rule may update non-decisive metadata. This is a point-in-time
property. It becomes a commit-time safety property only for required evidence
covered by $B(e)$; it does not assert that the agent chose the optimal action.

\subsection{Version and Decision Conflicts}

A \emph{version conflict} exists when a resource version differs from the one
recorded in $R$. A \emph{decision conflict} exists when current evidence
falsifies or makes unverifiable a reachable premise required by a pending
intent. The refund cases in Table~\ref{tab:trace} make the distinction explicit.
All five changes advance a recorded version. A new limit of 90 preserves
$80\leq90$ and therefore returns \textsf{KEEP}; a limit of 50 falsifies the
delegated-limit premise, and an already-issued refund falsifies the
not-yet-refunded premise, so both return \textsf{BLOCK}. A display-name change
affects only presentation and invokes an explicit \textsf{REFRESH} rule. A
corrected defect flag is a typical \textsf{REPLAN} case: the current refund is
no longer justified, but the task may be reconsidered. A version conflict is
therefore a trigger for inspection, not by itself a reason to discard the
plan. Table~\ref{tab:outcomes} defines these responses.

\begin{table}[t]
\caption{Validation outcomes. Severity increases from top to bottom; dominance
is $\textsf{BLOCK}>\textsf{REPLAN}>\textsf{REFRESH}>\textsf{KEEP}$.}
\label{tab:outcomes}
\small
\begin{tabular}{@{}lp{5.3cm}@{}}
\toprule
\textbf{Outcome} & \textbf{Required runtime response} \\
\midrule
\textsf{KEEP} & Retain the current intent; no recorded premise is falsified and
no refresh-only rule fires. \\
\textsf{REFRESH} & An explicit rule refreshes non-decisive presentation or
derived metadata; no decision premise is false. \\
\textsf{REPLAN} & The current intent is no longer justified, although the task
may remain feasible. \\
\textsf{BLOCK} & Do not release the effect; a safety, authority, policy, or
irreversible-effect condition is false, missing, contradictory, or unverifiable. \\
\bottomrule
\end{tabular}
\end{table}

\begin{table}[t]
\caption{Post-decision changes for the proposed 80 GBP refund. Every row
changes a connector-local version.}
\label{tab:trace}
\footnotesize
\begin{tabular}{@{}>{\raggedright\arraybackslash}p{2.35cm}
>{\raggedright\arraybackslash}p{2.45cm}l@{}}
\toprule
\textbf{Change after decision} & \textbf{Required condition} & \textbf{Outcome} \\
\midrule
Display name changes & No required premise changes & \textsf{REFRESH} \\
Limit: $100\rightarrow90$ & $80\leq90$ remains true & \textsf{KEEP} \\
Defect: true $\rightarrow$ false & Defect premise becomes false & \textsf{REPLAN} \\
Limit: $100\rightarrow50$ & Delegated-limit premise is false & \textsf{BLOCK} \\
Another worker refunds & Not-yet-refunded premise is false & \textsf{BLOCK} \\
\bottomrule
\end{tabular}
\end{table}

\subsection{Conditional Safety and Localization}

ATR maintains four invariants: every gated effect references its intent and
supporting decisions; every decision has typed premises or conservative edges
from every supplied input; every validation record names the evidence and
validator versions checked; and explanations are append-only. These invariants
support the following conditional property.

\textbf{Proposition 1 (bound pre-effect safety).} Assume (i) dependency
completeness: every evidence item capable of affecting $e$ has a recorded path
to $e$; (ii) each validator is sound for its declared typed premise; (iii)
token faithfulness: any change capable of altering a bound premise changes its
connector version or token; (iv) observation binding: the token supplied to
the target is the exact token paired with the observation evaluated by the
validator; and (v) atomic target enforcement: the target accepts $e$ only
while every required bound token in $B(e)$ still matches. Then an effect with
a false block-critical premise covered by $B(e)$ cannot commit through the ATR
gate.

\emph{Proof sketch.} Dependency completeness places the changed evidence and
supporting decision in the affected closure. Validator soundness maps a false
block-critical premise to \textsf{BLOCK}, which dominates the other outcomes,
so the gate does not invoke the effect. If a bound premise changes after
validation, token faithfulness and observation binding make the atomic target
condition fail; the rejected apply is not a commit and returns to validation.
A target-only CAS does not protect an external policy, budget, or authorization
resource whose token is absent from $B(e)$. The proposition therefore does not
claim global serializability, cross-system atomicity, exactly-once effects, or
complete recovery, and it does not cover an omitted premise or unsound
validator.

\textbf{Proposition 2 (graph-relative localization).} With exact reverse-index
lookups, affected closure computation visits only nodes reachable from matching
reads, in time $O(|\mathcal{A}_c|+|V_c|+L)$ for lookup cost $L$ and visited
subgraph $(V_c,\mathcal{A}_c)$. The traversal is confined to the subgraph
reachable under the recorded dependency relation; conservative edges may
include irrelevant nodes, and missing edges are unsafe.

\section{Runtime Design}

\subsection{Capture and Change Localization}

Connectors return resource identity, current value, version descriptor, and
conditional-effect capability. Versions may be ordered counters, commit
timestamps, ETags, hashes, or equality-only tokens. ATR records the actual
observation; monotonic or compatible caching can improve the read path but
does not determine whether a pending decision survives a later change. MCCache,
for example, enforces monotonic and consistent cache observations for
transaction-like requests~\cite{an2023mccache}; ATR instead links observations
to decisions and effects.

Dependencies can be captured conservatively for every input supplied to a
reasoning step, explicitly by application code, or from structured model
output. Reverse-index entries may select a whole decision or premise
identifiers attached to it. Object/field capture revisits every premise on a
selected decision; predicate capture can revisit only explicitly indexed
premises. Thus predicate capture may run one validator where coarser capture
runs all validators on that decision.

Free-form rationale cannot authorize an effect. When required typed premises
are known but dependency attribution is uncertain, ATR may link every supplied
input conservatively. This avoids missing-edge false allows but cannot repair
an omitted or unsound premise. A changed conservatively linked input without a
precise validator maps to \textsf{REPLAN} or \textsf{BLOCK}; authority- or
irreversible-effect-critical cases fail closed.

\begin{algorithm}[t]
\caption{Affected closure validation. Status ordering is
$\textsf{BLOCK}>\textsf{REPLAN}>\textsf{REFRESH}>\textsf{KEEP}$.}
\label{alg:validation}
\footnotesize
\begin{algorithmic}[1]
\STATE \textbf{procedure} \textnormal{Validate}$(x,e)$
\STATE $\Delta \gets \textnormal{normalize\_event}(x)$; $\mathit{obs}\gets\emptyset$
\IF{$x=\textsf{PRE\_EFFECT}$}
  \STATE $\mathit{obs}\gets\textnormal{read\_current}(\textnormal{required\_evidence}(e))$
  \STATE $\Delta\gets\Delta\cup\textnormal{version\_diff}(e,\mathit{obs})$
\ENDIF
\STATE $C_{\mathrm{aff}}\gets\textnormal{reachable\_pending}(\textnormal{index}(\Delta))$
\FORALL{$p \in \textnormal{required\_premises}(C_{\mathrm{aff}})$}
  \IF{$X(p)\nsubseteq\operatorname{dom}(\mathit{obs})$}
    \STATE $\mathit{obs}\gets\mathit{obs}\cup\textnormal{read\_current}(X(p)\setminus\operatorname{dom}(\mathit{obs}))$
  \ENDIF
  \STATE $\sigma[p]\gets\textnormal{validate}(p,\textnormal{observations\_for}(X(p),\mathit{obs}))$
\ENDFOR
\STATE $o\gets\textnormal{highest\_status}(\sigma,\textnormal{refresh\_rules}(\Delta))$
\STATE $(o_b,b)\gets\textnormal{binding\_policy}(e,\mathit{obs},B(e))$; $o\gets\max(o,o_b)$
\STATE $a\gets\textsf{NOT\_ATTEMPTED}$
\IF{$x=\textsf{PRE\_EFFECT}\ \wedge\ \textnormal{permits}(o,e)$}
  \STATE $a\gets\textnormal{conditional\_apply}(e,b)$
  \IF{$a=\textsf{REJECTED}$}
    \STATE \textnormal{schedule\_revalidation}$(e)$
  \ENDIF
\ENDIF
\STATE \textnormal{append\_explanation}$(x,C_{\mathrm{aff}},\sigma,o,b,a)$
\STATE \textbf{return} $(o,a)$
\end{algorithmic}
\Description{Pseudocode combines event hints with authoritative pre-effect
version comparisons, traverses the dependency graph, evaluates reachable
premises against value--token observations, records binding tokens, and only
then conditionally applies the effect.}
\end{algorithm}

Change streams and webhooks are early-warning optimizations, not the
correctness path. \textsc{normalize\_event} converts a hint into changed
resource/field pairs $\Delta$; \textsc{version\_diff} compares every required
evidence token at \textsf{PRE\_EFFECT}, including \textsf{REPLAN} premises.
Each connector returns a value and token as one observation, so $b$ contains
the exact tokens paired with the values validated. \textsc{refresh\_rules}
contains explicit application rules for non-decisive metadata.
\textsc{binding\_policy} returns \textsf{BLOCK} for unbound authority, safety,
or irreversible-effect evidence; any lower-risk relaxation must be explicit.
Object-only mismatches conservatively seed the resource. A missing or timed-out
validator follows the same \textsf{REPLAN}/\textsf{BLOCK} policy. Apply status
$a$ remains separate from the four validation outcomes; rejection never means
that a \textsf{KEEP} effect committed.

\subsection{Closing the Validation--Effect Race}

Validation alone leaves a time-of-check/time-of-use gap. For target-local
premises, ATR passes the checked target version into a native transaction,
CAS, uniqueness constraint, lease, or domain command. If the condition fails,
ATR revalidates or blocks. This does not by itself close a race on an external
policy, budget, or authorization record: such a change need not advance the
target version. Strong binding requires the target transaction to check those
versions, or an equivalent reservation, capability, lease, or fencing token.
Otherwise ATR records the evidence as unbound and exposes the residual race to
policy; an irreversible effect should fail closed. Concurrent agents need not
serialize their full reasoning, but the guarantee is limited to the tokens the
authoritative target actually checks.

ATR is not a workflow or recovery protocol. Sagas address long-running recovery
through compensating transactions~\cite{garciamolina1987sagas}; effect-staging
systems can suppress or compensate tool calls. ATR can supply them with
decision provenance but does not solve crashes between an external effect and
its acknowledgement, partial multi-system effects, or non-idempotent APIs.

\section{Prototype and Methodology}

We implemented a compact ATR runtime core in Python 3.11. The prototype uses immutable
read/decision/intent records, SHA-256 value digests,
hash-based reverse indexes, typed Python validators, append-only JSON records,
and an in-memory target exposing a version-token unit check. SQLite harnesses
exercise two-client target races, durable checkpoint/resume, and
cross-resource commit conditions. The runtime supports object, field, and
predicate granularity. The serialized graph averaged \GraphBytes\,KiB per trace with 48
extra context reads; this is an uncompressed prototype representation, not a
storage lower bound.

The evaluation asks: \textbf{RQ1}, does ATR reproduce the declared semantic
outcome and prevent invalid effects? \textbf{RQ2}, what selectivity and runtime
cost follow from reverse traversal? \textbf{RQ3}, where do races, unavailable
validators, and incomplete dependencies break the guarantee?

\textbf{Workloads.} We built three deterministic, executable scenarios:
refund (delivery, defect, duplicate-refund, and delegated-limit premises),
cloud remediation (incident state, duplicate remediation, restart authority,
and maintenance freeze), and procurement (request state, approved vendor,
budget, auto-approval cap, and purchasing authority). Each scenario has five
mutations: one irrelevant (\textsf{KEEP}), one presentation-only
(\textsf{REFRESH}), one decision-invalidating (\textsf{REPLAN}), and two
authority/target-critical (\textsf{BLOCK}). Thus, 40\% of the fixed mutation mix
is safe for the existing intent and 60\% is not. The oracle is the same explicit
business rule used to define each mutation category; no LLM labels results.

\textbf{Durable workflow cells.} Following CoAgent's evaluation pattern of
adding contending peers to existing tasks~\cite{lyu2026coagent}, we construct
one office-automation pattern and one cloud-remediation pattern, inspired by
task patterns represented in WorkBench~\cite{styles2024workbench} and
AIOpsLab~\cite{chen2025aiopslab}. For each, five intervening changes from a
second actor comprise an irrelevant update, a presentation-only update, a
change that preserves a predicate, a replanning case, and a blocking case. The workflow checkpoints to SQLite, closes its
connection, a second connection changes shared state, and a fresh connection
rebuilds the validation graph from checkpoint state and the versioned validator
registry before gating the effect. Ten trials per cell and method yield 400
method executions (10 cells $\times$ 10 repetitions $\times$ 4 methods). These
deterministic patterns do not run an LLM and are not full
upstream benchmark suites; they test persistence and target interaction beyond
the in-memory core.

\textbf{Baselines and metrics.} \emph{NoCheck} releases the original intent.
\emph{VersionAbort} replans on every version mismatch. \emph{FullScan} scans
all recorded provenance without a reverse index and re-evaluates every typed
premise; it does not replay the LLM or workflow. We report invalid effects,
false allows/blocks against the oracle, unnecessary replans, closure size,
serialized bytes, and in-process validation latency. Main and scaling latency
measure the warmed in-process validation call, including traversal and typed
premise validators,
but excluding graph construction, serialization, logging, checkpoint I/O,
connector/network calls, model calls, and replanning. The separate durable-cell
timer includes SQLite connect/read, JSON restoration, graph reconstruction,
validation, and the conditional update; those timings are not plotted or used
in Table~\ref{tab:primary}.

We run \RepetitionCount\ repeated balanced schedules with distinct shuffle
seeds, 10,000 traces per workload per repetition
(\TraceCount\ executions), on 16 vCPUs of an Intel Xeon Gold 6266C with 62\,GiB RAM
and Linux 6.6. For each repetition we checkpoint raw trace rows. Scaling holds the
affected predicate constant while increasing unrelated recorded reads. We also
run \SQLitePairs\ two-client SQLite conditional-update races across seven
repetitions, fail-closed validator tests, and a dependency-omission
fault-injection study with 35,000 unsafe trials at each omission rate. Source,
commands, checkpoints, raw CSV, independent audit scripts, and summary JSON
are in the artifact.

For the binding boundary, we repeat one scheduled history 100 times per mode:
validate, change an external refund policy, then attempt the target effect.
The repetitions stabilize timing only; they are not distinct histories. The
target-only mode checks the order version, while the bound mode atomically
checks both the order and recorded policy token.

The \TraceCount\ executions repeat 15 semantic mutation cases (three workloads
times five mutations); they are not \TraceCount\ distinct scenarios. The
repetitions check deterministic outcome agreement and measure runtime
variation under shuffled schedules.

\begin{figure*}[t]
  \centering
  \includegraphics[width=\textwidth]{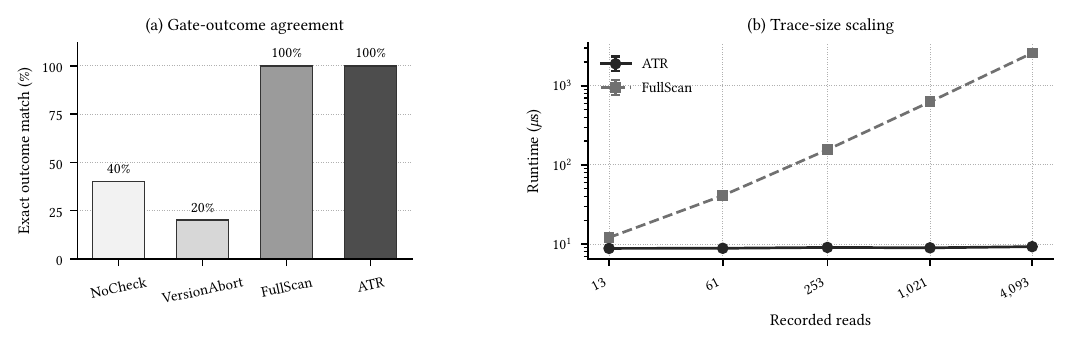}
  \caption{Gate-outcome agreement and trace-size scaling. (a) Exact four-way
  outcome agreement on ten controlled durable workflow cells. (b) In-process
  ATR and FullScan runtime versus recorded reads (log--log axes; 95\% CIs over
  seven repetitions).}
  \Description{Outcome-agreement bars show NoCheck at 40 percent,
  VersionAbort at 20 percent, and FullScan and ATR at 100 percent. A log-log
  plot shows ATR nearly flat and FullScan increasing with recorded reads.}
  \label{fig:results}
\end{figure*}

\section{Results}

\begin{table}[t]
\caption{Summary metrics. Unsafe-release and safe-replan rates are conditioned
on their respective controlled mutation classes; premises/change is measured
on durable workflow cells; p50 latency measures in-process validation.}
\label{tab:primary}
\footnotesize
\setlength{\tabcolsep}{2.6pt}
\begin{tabular}{@{}lrrrr@{}}
\toprule
\textbf{Method} & \shortstack{\textbf{Unsafe}\\\textbf{released (\%)}}
& \shortstack{\textbf{Safe}\\\textbf{replanned (\%)}}
& \shortstack{\textbf{Premises}\\\textbf{/ change}}
& \shortstack{\textbf{p50}\\\textbf{($\mu$s)}} \\
\midrule
NoCheck & \NoCheckUnsafeRelease & 0.0 & --- & --- \\
VersionAbort & 0.0 & \VersionAbortSafeReplan & --- & --- \\
FullScan & 0.0 & 0.0 & \FullScanPremisesPerChange & \FullScanMedian \\
ATR & 0.0 & 0.0 & \ATRPremisesPerChange & \ATRMedian \\
\bottomrule
\end{tabular}
\end{table}

\textbf{Controlled safety and selectivity.} Table~\ref{tab:primary} reports the
core result. NoCheck released 126,000 invalid effects (\NoCheckInvalid), a
direct consequence of the controlled mutation mix. VersionAbort released none
but replanned all 84,000 safe changes. ATR produced the oracle outcome for every
trace (\ATRFalseAllow\ false allows and \ATRFalseBlock\ false blocks), retaining
or refreshing those 84,000 intents. These observations demonstrate correct
implementation for the declared rules; they do not estimate real-world change
frequencies or premise-extraction accuracy.

\textbf{Durable workflow cells.} Figure~\ref{fig:results}(a) reports exact
outcomes for the local durable integrations. Across ten semantic cells, NoCheck and VersionAbort
matched the exact operational outcome in 40\% and 20\%, respectively;
FullScan and ATR matched all 100 executions per method. As summarized in
Table~\ref{tab:primary}, ATR evaluated \ATRPremisesPerChange\ typed premises per
resumed change, versus \FullScanPremisesPerChange\ for FullScan. These results establish
durable checkpoint/resume and SQLite interaction for the declared cells, not
agent reasoning quality.

\textbf{Cost and granularity.} ATR's p50/p95/p99 in-process validation latency is
\ATRMedian/\ATRPninetyfive/\ATRPninetynine\,$\mu$s; per-repetition p50 is
\ATRSeedMedianMean$\pm$\ATRSeedMedianCI\,$\mu$s (95\% CI). Predicate traversal visits
\PredicateClosure\ nodes on average (p95 3), versus \ObjectClosure\ (p95 8)
for whole-object capture. FullScan is \FullScanMedian\,$\mu$s at the workload trace size.
When unrelated reads increase from \SmallestReads\ to \LargestReads,
ATR changes from \SmallestSelective\ to \LargestSelective\,$\mu$s, whereas FullScan rises from
\SmallestFullScan\ to \LargestFullScan\,$\mu$s (Figure~\ref{fig:results}(b)). This is a
single-process microbenchmark; distributed-store and connector latency remain
to be measured.

\textbf{Races and failure modes.} Across \SQLitePairs\ pairs, two independent
SQLite connections were released at a barrier to issue the same atomic
version-conditional update. Exactly one update committed per pair
(\SQLiteDoubleCommits\ double-commit and \SQLiteZeroCommits\ zero-commit pairs).
This local-backend test validates the binding pattern, not distributed-database
behavior. In the deliberately scheduled cross-resource history, target-only
CAS committed 100/100 effects after the policy limit became false; adding the
checked policy token rejected 100/100. This is a counterexample and mechanism
check, not an estimated production failure rate. Three targeted unavailable-validator tests all
returned \textsf{BLOCK}. Randomly omitting the sole critical
dependency edge produced false-allow rates of 0.95\%, 5.05\%, and 9.97\% at
1\%, 5\%, and 10\% omission. Selective validation is safe only when dependency
capture is complete or conservatively over-approximated.

\section{Related Work}

The related systems address different layers of a long-running agent action.
Database isolation and OCC control read/write conflicts and atomic target
updates~\cite{berenson1995isolation,kung1981occ}; semantic serializability uses
application semantics to enlarge the set of correct database schedules
~\cite{brayner1999semantic}. ATCC adapts optimistic and pessimistic control to
long, unforeseen SQL transactions~\cite{zhou2026atcc}. These mechanisms govern
data execution; they do not determine whether a changed business fact
invalidates the reasoning behind a pending agent action.

At the agent-concurrency layer, CoAgent orders concurrent peers, reports
conflicts to LLM workers, and repairs or reverses writes to target
serializability~\cite{lyu2026coagent}. ATR neither orders agents nor repairs
visible writes; it checks the stated conditions for one pending action.
SemIso instead keeps prompts, models, indexes, policies, and tools compatible
across durable continuations~\cite{mozafari2026semiso}. It protects the
workflow's semantic environment, while ATR asks whether mutable business
evidence still supports a particular decision.

Other work addresses admission, effect release, or recovery. Mnemosyne admits
generated proposals under executable constraints and repairs disruptions
~\cite{chang2026mnemosyne}. Cordon validates a composed flow before releasing
staged effects~\cite{chen2026cordon}; Atomix uses progress frontiers to decide
when speculative or contending effects may settle
~\cite{mohammadi2026atomix}. Sagas provide compensation for long-running
recovery~\cite{garciamolina1987sagas}, and Agentic Transaction proposes a
broader semantic-ACID architecture~\cite{sun2026agentictransaction}. ATR
occupies the narrower decision-validation layer between observation and effect:
it asks whether changed evidence has falsified a condition that justified the
pending action, then relies on the target system for atomic commit.

\section{Limitations and Conclusion}

The evidence is deliberately narrow. ATR runs on one host; the task patterns
inspired by WorkBench/AIOpsLab are not complete upstream runs; workloads
are deterministic; premise validators and ground truth
are developer-authored; and no experiment measures natural-language premise
extraction, model replanning quality, distributed failure, or production
change distributions. A model-generated premise must be treated as untrusted
until mapped to an executable validator. When dependency attribution is
uncertain but the required typed premises are known, conservatively linking all
supplied inputs avoids missing-edge false allows at the cost of additional
revalidation. This does not compensate for an omitted or unsound premise.

Within those boundaries, ATR demonstrates that version mismatches need not
force an all-or-nothing choice between ignoring change and discarding all prior
reasoning. Recording evidence-to-effect dependencies, revisiting the affected
closure, and binding the result to checked target tokens can retain harmless
decisions while blocking invalid ones at microsecond-scale in-process
validation cost. For
cross-resource premises, that guarantee requires a transaction or explicit
reservation/lease mechanism covering the external evidence. The next step is integration with a durable agent framework and
transactional connectors. It should then be evaluated on production-derived
traces with measured premise-extraction error.

\section*{Generative-AI Use Disclosure}

OpenAI Codex was used for prototype and code scaffolding, experiment-script
review, figure generation, and language editing. Executable rules and
evaluation labels are developer-authored, and reported aggregates were
independently recomputed from raw checkpoints. The human authors inspected and
take responsibility for all submitted code, data, citations, and claims.

\section*{Artifacts}

The public artifact is available in the GitHub repository
\href{https://github.com/ezreal13/atr-decision-validation/tree/89c32fa475f5216db4c9d76a1eddde79e874073b}%
{\nolinkurl{ezreal13/atr-decision-validation}} (commit
\texttt{89c32fa}).
It contains the complete Python prototype, workload and mutation definitions,
tests, controlled and anchored checkpoints, raw CSV, SQLite databases and
outputs, binding traces, independent audit reports and scripts, summary JSON,
the figure generator and editable figure sources, and exact reproduction
commands. The experiment requires Python 3.11 and NumPy; it runs without
network access or model credentials. Each repetition can be resumed
independently.

\bibliographystyle{ACM-Reference-Format}
\bibliography{references}

\end{document}